\documentclass[runningheads]{llncs}
\usepackage[T1]{fontenc}
\usepackage{graphicx}

\usepackage{amsmath} 
\usepackage{multirow}
\usepackage{amssymb,amsfonts}
\begin{document}
\title{ProFi-Net: Prototype-based Feature Attention with Curriculum Augmentation for WiFi-based Gesture Recognition}
\titlerunning{ProFi-Net}
%
\author{Zhe Cui\inst{1} \and
Shuxian Zhang\inst{2} \and
Kangzhi Lou\inst{1} \and
Le-Nam~Tran\inst{1}}
%
\authorrunning{Cui et al.}
%
\institute{School of Electrical and Electronic Engineering, University College
Dublin, Ireland
\email{zhe.cui@ucdconnect.ie, kangzhi.lou@ucdconnect.ie nam.tran@ucd.ie}\\
\and
College of Electrical Engineering and Automation, Shandong University of Science and Technology, China\\
\email{202282080077@sdust.edu.cn}}


%
\maketitle              
\begin{abstract}
This paper presents ProFi-Net, a novel few-shot learning framework for WiFi-based gesture recognition that overcomes the challenges of limited training data and sparse feature representations. ProFi-Net employs a prototype-based metric learning architecture enhanced with a feature-level attention mechanism, which dynamically refines the Euclidean distance by emphasizing the most discriminative feature dimensions. Additionally, our approach introduces a curriculum-inspired data augmentation strategy exclusively on the query set. By progressively incorporating Gaussian noise of increasing magnitude, the model is exposed to a broader range of challenging variations, thereby improving its generalization and robustness to overfitting. Extensive experiments conducted across diverse real-world environments demonstrate that ProFi-Net significantly outperforms conventional prototype networks and other state-of-the-art few-shot learning methods in terms of classification accuracy and training efficiency.

\end{abstract}
\section{Introduction}


Gesture recognition is a key technology in Human-Computer Interaction (HCI) with broad applications in smart homes, healthcare, autonomous systems, and assistive technologies \cite{hao2020wi}. For example, in smart homes, it enables touchless control of devices like air conditioners and lighting. In healthcare, it supports contactless patient monitoring and rehabilitation, while in autonomous driving, it enhances driver assistance through hand gestures. Moreover, it plays a crucial role in sign language recognition, motion tracking, and object detection \cite{liu2020human}, improving accessibility, safety, and user experience across various domains.

Conventional gesture recognition mainly relies on wearable sensor-based and computer vision-based approaches.
Wearable sensor-based gesture recognition \cite{botros2022day,qi2020surface,shen2020gesture} detects motion patterns by capturing position, speed, and direction through specialized sensors. However, it requires users to wear specialized devices, making it inconvenient for daily use.
Computer vision-based gesture recognition \cite{hussain2021comprehensive,gao2020vision} extracts gesture features from images or videos. Although sensor-free and convenient, it has drawbacks, including sensitivity to lighting conditions, high computational costs, and privacy concerns.

Unlike these methods, wireless-based gesture recognition offers a device-free solution, eliminating the need for wearable sensors and remaining unaffected by lighting conditions \cite{peng2025stability,anguita2025multi}. It recognizes gestures by capturing gesture-induced variations in wireless signals. Among wireless-based approaches, continuous wave radar and ultra-wideband (UWB) systems achieve high accuracy but rely on specialized, costly hardware, limiting their feasibility for widespread deployment.

With the ubiquitous deployment of WiFi networks, WiFi-based gesture recognition \cite{chen2019dynamic,gu2023wireless} has emerged as a promising solution, offering passive sensing, low-cost hardware, and easy deployment while overcoming the limitations of sensor-based and vision-based methods. 
The development of the CSI-Tool, which extracts Channel State Information (CSI) from wireless network cards, has further enabled fine-grained gesture recognition through richer subcarrier amplitude and phase information in WiFi signals. As a result, a growing number of studies have explored WiFi-based gesture recognition.

Traditional machine learning methods have been widely used in WiFi-based gesture recognition. Sruthi et al. \cite{sruthi2024handfi} proposed a WiFi-based gesture recognition model that classifies gestures by analyzing CSI variations, applying K-Nearest Neighbors (KNN), Decision Trees, and Support Vector Machine (SVM) for classification. 
Similarly, Tian et al. \cite{tian2018wicatch} introduced WiCatch, a device-free WiFi-based gesture recognition system leveraging leveraging CSI and employing SVM for classification.

Compared to traditional machine learning, deep learning has been widely adopted due to its powerful feature extraction capabilities.
Li et al. \cite{li2020wihf} proposed WiHF, a WiFi-based system for simultaneous gesture recognition and user identification, utilizing a deep neural network (DNN) with splitting and splicing schemes for optimized collaborative learning.
Wang et al. \cite{wang2018channel} introduced CSAR, a WiFi-based activity recognition system that employs Long Short-Term Memory (LSTM) networks for enhanced accuracy and robustness. CSAR dynamically selects high-quality WiFi channels, achieving 95\% accuracy.
Meng et al. \cite{meng2021wihgr} proposed WiHGR, a WiFi-based gesture recognition system addressing model complexity and accuracy issues in dynamic environments. It utilizes a modified attention-based bi-directional gated recurrent unit (ABGRU) network to extract discriminative features, with an attention mechanism assigning higher weights to critical features.
Bu et al. \cite{bu2022deep} introduced a WiFi-based gesture recognition method using deep transfer learning. CSI streams are captured, segmented, and transformed into an image matrix, with Convolutional Neural Network (CNN)-based transfer learning techniques extracting high-level features for recognition.
Tang et al. \cite{tang2021wifi} proposed deep space-time neural networks, specifically a Long Short-Term Memory-Fully Convolutional Network (LSTM-FCN), for spatio-temporal feature extraction in CSI-based gesture recognition.
Kabir et al. \cite{kabir2022csi} introduced CSI-DeepNet, a lightweight deep learning-based gesture recognition system for resource-limited devices. Unlike conventional CNN-based methods that require high computational resources, CSI-DeepNet reduces complexity while maintaining high accuracy.

Despite their high accuracy, deep learning-based WiFi gesture recognition methods heavily rely on large-scale labeled datasets. However, in practical applications, collecting extensive labeled WiFi CSI gesture data is costly, time-consuming, and often impractical, especially for rare or complex gestures.  Additionally, privacy concerns, high data collection costs, and ethical constraints further restrict access to high-quality labeled gesture data, making it challenging to develop and deploy WiFi-based gesture recognition systems.
To address the issue of limited training samples, few-shot learning (FSL) has emerged as a promising solution, allowing models to generalize to new gesture categories with only a few labeled examples. However, existing few-shot learning methods inherently lacks sufficient samples, leading to sparse feature representations, making it difficult for the model to learn discriminative features for WiFi-based gesture recognition.

To overcome these limitations, this paper proposes ProFi-Net (Prototype-based Feature Attention with Curriculum Augmentation for WiFi-based Gesture Recognition), a novel few-shot learning network that enables high-accuracy recognition of new gesture categories with minimal labeled samples. ProFi-Net integrates feature attention-enhanced metric learning with curriculum-based data augmentation (CDA) to improve model generalization and classification accuracy. Specifically, a feature attention-based prototype network is designed to enhance feature discrimination and mitigate the effects of feature sparsity in few-shot learning. Additionally, CDA gradually increases the complexity of training samples, allowing the model to adapt more effectively to gesture variations, reduce overfitting, and improve recognition performance. The key contributions of this work are summarized as follows:

\begin{itemize}
    \item We propose ProFi-Net, a novel few-shot learning model designed to tackle the data inefficiency in deep learning-based gesture recognition, enabling accurate classification of new gestures with only a few labeled samples.
    To address feature sparsity in few-shot learning, we introduce a feature-level attention mechanism into the distance metric module of the prototype network, improving feature discrimination and classification performance.
    \item A curriculum-based data augmentation approach is employed, where training begins with simpler samples and gradually introduces more complex variations, improving generalization and robustness against overfitting.
    \item Extensive experiments demonstrate that ProFi-Net outperforms state-of-the-art few-shot learning models for WiFi-based gesture recognition, achieving high classification accuracy while significantly reducing data requirements.
\end{itemize}

In the following, we first introduce the overall architecture of ProFi-Net in Section~\ref{Problem Setting}, followed by a detailed description of its key components in Section~\ref{Methodology}. Specifically, we present the representation learning module, the prototype-based metric learning with feature-level attention, and the curriculum-inspired data augmentation strategy. Section~\ref{Experiment and Evalutions} details our experimental setup and evaluation results, and finally, Section~\ref{Conclusion} concludes the paper with discussions on the outcomes and future research directions.

\section{Problem Setting}
\label{Problem Setting}

WiFi-based Few-Shot Learning for Gesture Recognition aims to classify previously unseen gestures using only a limited number of labeled examples. Unlike conventional deep learning methods that rely on large-scale labeled datasets, few-shot learning leverages meta-learning techniques to generalize from sparse data. Let $\mathcal{X}$ denote the CSI feature space and $\mathcal{Y}$ the gesture label space. Each WiFi gesture instance is represented as $(x, y)$, where $x \in \mathcal{X}$ corresponds to the CSI measurements and $y \in \mathcal{Y}$ is the associated gesture label.

The few-shot learning dataset is partitioned into two disjoint subsets, denoted as $\mathcal{D} = (\mathcal{D}_{trn}, \mathcal{D}_{tst})$. The meta-training phase employs $\mathcal{D}_{trn}$ to learn robust feature representations and develop the capacity to quickly adapt to new tasks, while the meta-testing phase assesses the model's generalization on unseen gesture categories using $\mathcal{D}_{tst}$. Accordingly, the label sets $\mathcal{Y}_{trn}$ and $\mathcal{Y}_{tst}$ are mutually exclusive (i.e., $\mathcal{Y}_{trn} \cap \mathcal{Y}_{tst} = \emptyset$), ensuring that gesture classes encountered during training do not appear during testing.

Few-shot learning is structured as a series of episodic tasks, each adhering to an $N$-way $K$-shot classification setting. In each episode, $N$ gesture classes are randomly selected. For each selected class, $K$ labeled examples are sampled to form the Support Set $\mathcal{S} = \{(x_s^i, y_s^i)\}_{i=1}^{N \times K}$, $x_s^i \in \mathcal{X}$ represents the CSI measurement and $y_s^i \in \mathcal{Y}_{trn}$ (or $\mathcal{Y}_{tst}$, depending on the phase) is its corresponding label. Additionally, a Query Set $\mathcal{Q} = \{(x_q^j, y_q^j)\}_{j=1}^{M}$ is constructed by sampling $M$ examples from the same $N$ classes, $x_q^j \in \mathcal{X}$ denotes the CSI measurement of the $j$-th query sample, and $y_q^j$ is its ground-truth label.

During meta-training, the model is exposed to numerous episodic tasks sampled from $\mathcal{D}_{trn}$, which refines its ability to rapidly adapt to new classification challenges. In the meta-testing phase, the model is evaluated on similar $N$-way $K$-shot tasks derived from $\mathcal{D}_{tst}$, thereby assessing its capability to generalize to entirely novel gesture categories.

The objective of few-shot WiFi gesture recognition is to predict the class label $\hat{y}_q$ for each query sample $x_q$ based solely on the limited labeled examples in the Support Set during testing. By training on diverse episodic tasks, the model progressively enhances its capacity to recognize gestures with minimal labeled data, enabling efficient and scalable WiFi-based gesture recognition.

\section{Methodology}
\label{Methodology}
\subsection{System Overview}

\begin{figure}
\centering
\includegraphics[width=0.8\textwidth]{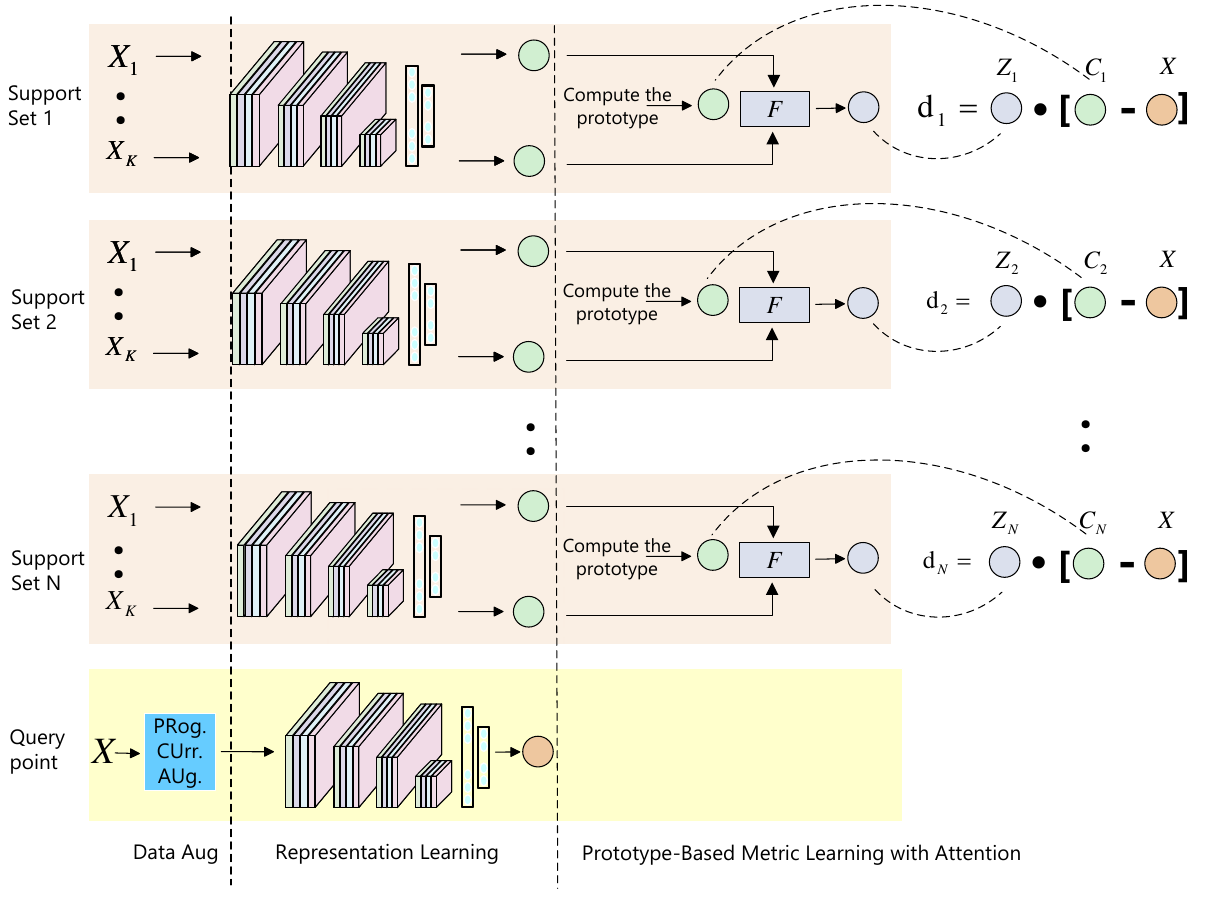}
\caption{The framework of the proposed ProFi-Net.} 
\label{fig1}
\end{figure}

ProFi-Net addresses WiFi-based few-shot gesture recognition by integrating metric learning with curriculum-based query augmentation. Unlike conventional deep learning models that require extensive labeled datasets, ProFi-Net is designed to learn effectively from limited samples and generalize to unseen gesture categories. As illustrated in Fig. \ref{fig1}, the overall framework comprises three main stages: Representation Learning, Prototype-Based Metric Learning with Attention, and Curriculum-Guided Query Augmentation.

To support few-shot learning, ProFi-Net is trained on episodic tasks. In each episode, a support set $\mathcal{S} = \{(x_s^i, y_s^i)\}$ and a query set $\mathcal{Q} = \{(x_q^j, y_q^j)\}$ are sampled from $N$ gesture classes, forming an $N$-way $K$-shot classification problem. For each gesture class, $K$ labeled samples are selected for the support set, while additional unlabeled samples are used to construct the query set.
Firstly, in the Representation Learning module, both support and query samples are processed through a four-layer convolutional neural network that learns a mapping function $f_{\phi}: \mathcal{X} \rightarrow \mathbb{R}^d$, $\phi$ denotes the learnable parameters, $\mathbb{R}^d$ represents $d$-dimensional embedding space. This mapping projects WiFi CSI signals into a high-dimensional embedding (or metric) space where samples from the same gesture class are closely clustered, aided by the non-linearity introduced by ReLU activations.
Secondly, once the feature vectors for the support set have been extracted, the system computes class prototypes by averaging the vectors corresponding to each gesture class. Simultaneously, these feature vectors are passed through a feature-level attention module $F$, which produces an attention score vector to emphasize the most discriminative feature dimensions. The query set is then mapped into the same embedding space, and an attention-based metric is used to compute the distance between each query sample and every class prototype. 
Thirdly, to further mitigate overfitting and enhance generalization in the few-shot setting, ProFi-Net employs a curriculum learning-based data augmentation strategy on the query set. Instead of conventional augmentation techniques such as rotation, translation, or cropping, this strategy gradually introduces varying levels of Gaussian noise into the query samples. By progressively increasing the noise level, the model is exposed to a wider range of sample difficulties, thereby promoting robust learning and ultimately improving classification accuracy.

\subsection{Data Preprocessing and Representation Learning}

Raw CSI data collected from WiFi devices typically contain significant noise that adversely affects recognition accuracy. To enhance the precision of the extracted CSI features, we apply a two-step denoising procedure: first, Hampel Filtering detects and removes outliers while preserving the structural integrity of the CSI signal; second, the Discrete Wavelet Transform (DWT) smooths the waveform and effectively reduces high-frequency noise components.

To capture fine-grained variations in CSI signals, we employ a four-layer convolutional neural network for representation learning. This CNN transforms raw CSI data into compact, discriminative embeddings. The network consists of four convolutional layers, each followed by batch normalization, ReLU activation, and max pooling, ensuring stable training and effective feature transformation. 

Let \( X_s \) and \( X_q \) represent the CSI data from the support set and query set, respectively, where 
\( X_s = \{ x_s^i \}_{i=1}^{K \times N} \) consists of \( K \) labeled examples per class across \( N \) classes, and 
\( X_q = \{ x_q^j \}_{j=1}^{M} \) contains \( M \) unlabeled query examples. 
The CNN extracts feature representations by applying a mapping function \( f_\phi(\cdot) \) parameterized by \( \phi \), 
transforming the raw CSI inputs into a discriminative feature space:

\[
    Z_s = f_\phi(X_s) = \{ z_s^i \}_{i=1}^{K \times N}, \quad 
    Z_q = f_\phi(X_q) = \{ z_q^j \}_{j=1}^{M}
\]
where \( Z_s \) and \( Z_q \) are the corresponding feature embeddings for the support and query samples.


\subsection{Prototype-Based Metric Learning with Attention}

The prototype network classifies query samples by comparing their embeddings to class prototypes, which are computed by averaging the representative vectors of all support samples within the same category. 
Given the support set embeddings $Z_s = \{ z_s^i \}_{i=1}^{K \times N}$, the prototype for class \( c \) is computed as:

\[
   C_c = \frac{1}{|S^c|} \sum_{(x_s^i, y_s^i) \in S^c} z_s^i
\]
where \( C_c \in \mathbb{R}^d \) represents the prototype of class \( c \), \( |S^c| \) is the number of support samples belonging to class \( c \). 

\begin{figure}
\centering
\includegraphics[width=0.4\textwidth]{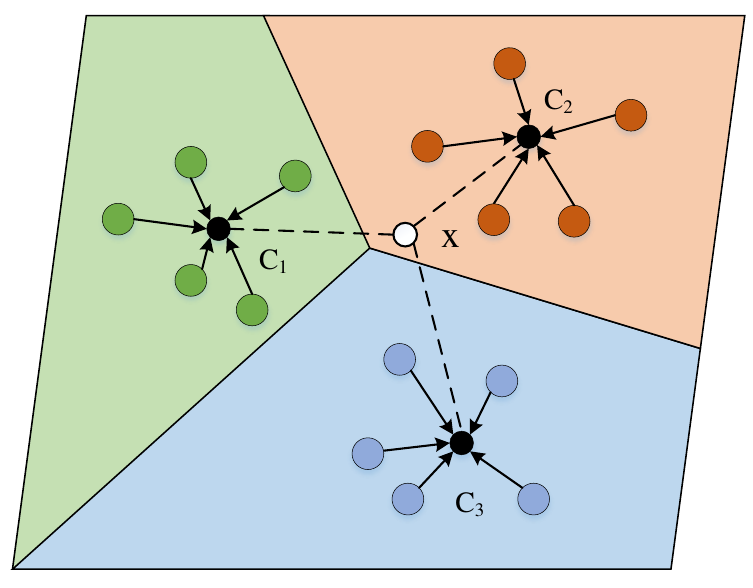}
\caption{Prototype representation in prototype network.} \label{fig:prototype_module}
\end{figure}

The prototype representation module is illustrated in Fig.~\ref{fig:prototype_module}.
As shown, \( C_1 \), \( C_2 \), and \( C_3 \) represent the prototypes of three different categories. For a given query point \( x \), classification is performed by finding the prototype closest to it. In this example, since \( x \) is closest to \( C_2 \), it is predicted to belong to class \( C_2 \).

After obtaining the prototypes, query samples are mapped into the same embedding space using \( f_{\phi}(\cdot) \), yielding embeddings \( Z_q = \{ z_q^j \}_{j=1}^{M} \) for the query set. The standard Euclidean distance between a query embedding \( z_q \) and a class prototype \( C_c \)  is given by:
\[
    d_{\text{Euc}}(x_q, C_c) = \| z_q - C_c \|^2.
\]

Due to the limited number of support samples, the extracted features may be sparse and might not fully capture the most discriminative aspects needed for classification. To address this, a feature-level attention mechanism \(F\) is introduced. This module generates an attention score vector \( A_c \) that emphasizes the most informative feature dimensions for class $c$. The distance function is then refined by incorporating these attention scores. Specifically, the modified distance function \( d(\cdot) \) is defined as:
\[
    d(x_q, C_c) = A_c \cdot \| z_q - C_c \|^2,
\]
where $A_c$ represents the learned attention scores specific to class $c$ . 
The computed distances are subsequently used in a softmax function to predict the class probabilities of the query samples:
\[
    p_{\phi}(y = c \mid x) = \frac{\exp\bigl(-d(f_{\phi}(x), C_c)\bigr)}{\sum_{c'} \exp\bigl(-d(f_{\phi}(x), C_{c'})\bigr)},
\]
where \( p_{\phi}(y = c \mid x) \) denotes the probability that query sample \(x\) belongs to class \(c\).

Finally, the model is trained by minimizing the negative log-likelihood loss:
\[
    J(\phi) = -\log p_{\phi}(y \mid x),
\]
where \(y\) represents the true label of the training sample. Optimization is performed using stochastic gradient descent (SGD) to iteratively adjust the model parameters, thereby promoting intra-class compactness and inter-class separation.

The structure of the feature-level attention module consists of three convolutional blocks, each comprising a convolutional layer followed by a ReLU activation. The inputs \( X_1, \dots, X_K \) (representing the \(K\) samples from each class) are processed through these blocks to generate a class-specific score vector. 

This enhanced prototype-based metric learning approach, incorporating a feature-level attention mechanism, effectively mitigates feature sparsity and improves the discriminative power of the learned embedding space.

\subsection{Data augmentation based on curriculum learning}

Few-shot classification tasks suffer from limited data, making models prone to overfitting. Although conventional augmentation techniques (e.g., rotation, flipping, cropping, and padding) are often used to enrich datasets, their impact on classification accuracy is generally limited. In our work, we adopt a curriculum learning-based data augmentation strategy, applied exclusively to the query set $\mathcal{Q}$ , to further enhance model performance.
The key idea is to gradually introduce augmented samples of increasing difficulty, allowing the model to progressively adapt to more challenging variations and thus improve its generalization capability. Unlike standard augmentation approaches, which treat all augmented data equally, our curriculum-based method assigns different weights to samples based on their difficulty. Initially, the model is trained solely on the original query data until a certain convergence level is reached. Subsequently, augmented samples—perturbed by Gaussian noise of increasing magnitude—are progressively introduced into the training set.

Specifically, our progressive curriculum augmentation on the query set is implemented in multiple stages. In the first stage, the model is trained on the original query data. In the second stage, query data is augmented with 10\% Gaussian noise; in the third stage, 20\% noise is added; and this process continues until the sixth stage, where the noise level reaches 50\%. Throughout all stages, the augmented data is combined with the original data at a fixed ratio of 4:1.

Gaussian noise is used as the perturbation, with its probability density function defined as follows:
\begin{equation}
    p(x) = \frac{1}{\sqrt{2\pi}\sigma} \exp\left(-\frac{(x-\mu)^2}{2\sigma^2}\right),
\end{equation}
where \(\mu\) and \(\sigma\) denote the mean and standard deviation, respectively.

We quantify the noise level using the Signal-to-Noise Ratio (SNR), which is defined in decibels (dB) as:
\begin{equation}
    \text{SNR (dB)} = 10 \log_{10} \left(\frac{P_{\text{signal}}}{P_{\text{noise}}}\right).
\end{equation}
According to the calculations, adding 10\% noise (i.e., noise level of 0.1) results in an SNR of approximately 20 dB, while 20\% noise (0.2) corresponds to an SNR of around 14 dB. Similar computations yield the SNR values for the other noise levels, which serve as the basis for our curriculum learning strategy.

\section{Experiment and Evalutions}
\label{Experiment and Evalutions}
\subsection{Experimental settings}
To evaluate the proposed model, we designed a prototype system to collect WiFi signals for predicting gesture. 
To record the WiFi signals, the prototype system employed two DELL laptops acting as the transmitter (TX) and the receiver (RX), both equipped with Intel 5300 network cards and the Linux 802.11n CSI Tool.
The TX and RX were mounted on tripods at a height of 1.2 meters, positioned 1.5 meters apart, facing each other.
To ensure the collection of fine-grained information about crowd counting, the sampling rate was set to 500 Hz, and each sample collection time window was set to 4 seconds. 
Data collection was conducted in three different environments: a spacious meeting room (Environment A), a cluttered laboratory (Environment B), and a semi-enclosed corridor (Environment C). 

The three environments exhibit significant differences. Environment A has an area of 84 m² (12m $\times $ 7m), with an open layout containing only wooden office desks and chairs, and occasional personnel entering and exiting during data collection. Environment B covers 76.5 m² (9m $\times $ 8.5m), and features a complex indoor setting with desks, chairs, various experimental equipment, and miscellaneous items. Personnel occasionally pass through during data collection. Environment C is a semi-enclosed corridor, with an area of 56 m² (8m $\times $ 7m), furnished with a sofa and a long table, with pedestrians occasionally passing by during data collection. 



The six volunteers with significant differences were asked to sit between the transmitter and receiver on a bench and use their right-hand fingers to draw the numbers 0–9 and the 26 uppercase and lowercase English letters, for a total of 62 gesture categories. 
Each gesture category was collected 50 times, resulting in a total of 18,600 CSI data packets for each environment. 
Since CSI provides fine-grained information, it is sensitive to slight differences in both the environment and gestures. 
Additionally, the stroke order of these numbers and letters may vary between individuals. 
To improve gesture recognition accuracy, standard gestures for these 62 categories were defined. 
Before data collection, the experimenter sat with both hands resting on their lap, arms naturally hanging down. 
The data collection started with an audio signal saying "Start," and the experimenter began performing gestures in the air. 
The complete gesture was made before the audio signal said "End." The experimenter was instructed to maintain a uniform speed throughout the movement, with brief pauses before and after each gesture. 
Each data collection cycle lasted 4 seconds, with approximately 3.5 seconds for performing the gesture. 



A total of 62 gesture categories were collected, each containing 50 samples. To maintain an integer class distribution, 46 categories were allocated for training and 16 for testing, following the N-way K-shot configuration used in few-shot learning.
The model was trained for 600 epochs, each consisting of 100 episodes, totaling 60,000 episodes. Noise levels were progressively increased at intervals of 100 epochs to assess model robustness, and the final performance was obtained by averaging results across all epochs. 
Training was conducted using the Adam optimizer with an initial learning rate of $1.0 \times 10^{-4}$.

For the N-way K-shot task, the number of query samples per class (\(N_q\)) was fixed at 10, following standard few-shot learning settings. Experiments were conducted under 5-way 1-shot and 5-way 5-shot scenarios. In each episode, 5 categories were sampled to form a 5-way task, with each category including \( k \) labeled samples in the support set (\( k = 1 \) for 1-shot, \( k = 5 \) for 5-shot). 
The same configuration was applied during testing.

To evaluate the effectiveness of the proposed method, we compare it with various few shot metric learning models, including MatchingNet (MN) \cite{vinyals2016matching}, RelationNet (RN) \cite{sung2018learning}, DeepEMD \cite{zhang2020deepemd}, FRN \cite{wertheimer2021few}, and DCAP \cite{he2022revisiting}. Experiments were conducted under 5-way 1-shot and 5-way 5-shot settings, with the number of query samples per class (\(N_q\)) fixed at 10. Accuracy and training time were used as evaluation metrics. To ensure fairness, all methods utilized a 4-layer CNN as the feature extractor and were trained for 600 epochs.

\subsection{Results and Discussions}
\subsubsection{Performance Overview}
Table \ref{tab:comparison_results} compares the performance of our proposed ProFi-Net with various other models, includes MatchingNet, RelationNet, DeepEMD, FRN and DCAP. 

\begin{table}[h]
\centering
\caption{Experimental results of comparative algorithms}
\label{tab:comparison_results}
\begin{tabular}{c c c c c c}
\hline
\multirow{2}{*}{Environment} & \multirow{2}{*}{Model} & \multicolumn{2}{c}{5-way 1-shot} & \multicolumn{2}{c}{5-way 5-shot} \\ \cline{3-6}
                             &                        & Accuracy (\%) & Training Time (s) & Accuracy (\%) & Training Time (s) \\ \hline
\multirow{6}{*}{A}           & MN                     & 51.0          & 87660.0           & 66.8          & 95653.8           \\ 
                             & RN                     & 58.9          & 45010.8           & 76.2          & 53438.1           \\ 
                             & DeepEMD                & 66.2          & 72480.9           & 82.4          & 80798.1           \\ 
                             & FRN                    & 64.4          & 41986.7           & 81.6          & 47582.6           \\ 
                             & DCAP                   & 66.0          & 51490.1           & 83.2          & 61880.2           \\ 
                                     & ProFi-Net,            & \textbf{66.5}          & \textbf{40547.4}           &\textbf{ 84.1}          & \textbf{49076.4}           \\ \hline
\multirow{6}{*}{B}           & MN                     & 45.5          & 84617.8           & 58.0          & 92601.0           \\ 
                             & RN                     & 52.5          & 43881.2           & 65.2          & 48329.1           \\ 
                             & DeepEMD                & 57.9          & 71391.2           & 72.0          & 78500.3           \\ 
                             & FRN                    & 55.8          & 40356.4           & 71.6          & 47019.2           \\ 
                             & DCAP                   & 58.0          & 49159.9           & 72.1          & 59852.3           \\ 
                                 & ProFi-Net            & \textbf{59.3}          & \textbf{38698.6}           &\textbf{ 72.5}          & \textbf{46180.9}           \\ \hline
\multirow{6}{*}{C}           & MN                     & 53.2          & 89773.8           & 68.0          & 97086.2           \\ 
                             & RN                     & 60.1          & 46250.5           & 74.9          & 54038.6           \\ 
                             & DeepEMD                & 67.5          & 74809.0           & 82.8          & 80513.6           \\ 
                             & FRN                    & 67.8          & 43015.4           & 83.9          & 48215.2           \\ 
                             & DCAP                   & 67.2          & 53936.5           & 84.1          & 63581.0           \\ 
                             & ProFi-Net            & \textbf{68.1}          & \textbf{42110.1}           & \textbf{85.7}          & \textbf{50547.9}           \\ \hline
\end{tabular}
\end{table}

Compared to other few-shot learning algorithms, the proposed ProFi-Net model demonstrates outstanding performance. In the 5-way scenario, it achieves high classification accuracy of 66.5\% (1-shot) and 84.1\% (5-shot), while also leading in model training efficiency. This indicates that the proposed model can quickly and effectively mitigate overfitting or weak generalization issues in few-shot learning. Furthermore, the results above show that the average accuracy in the 5-shot setting is approximately 17.5\% higher than in the 1-shot setting. This suggests that, similar to conventional deep learning, more training data leads to more learned information and features, which in turn increases classification accuracy. The accuracy of the MN network, which initially stood at 51.0\% and 66.8\% for 1-shot and 5-shot, has now been improved to approximately 66\% and 84\%, respectively.

\subsubsection{Ablation Study}
Since this model is an improved version of the ProtoNet algorithm, ProtoNet is used as the comparison network in this ablation study. Ablation experiments are conducted on ProtoNet with the feature-level attention mechanism ($A$), ProtoNet with fixed noise percentage (B), ProtoNet with progressive noise percentage ($B^{+}$), and ProtoNet with both progressive noise percentage and feature-level attention mechanism ($A+B^{+}$). The ablation experiment results in the three environments are shown in Table \ref{tab:ablation}. The experimental results use accuracy as the evaluation metric.

\begin{table}[ht]
\centering
\caption{Results of Ablation Experiments in Environments A, B, and C}
\label{tab:ablation}
\begin{tabular}{c|c|c|c}
\hline
\textbf{Environment} & \textbf{Model} & \textbf{5-Way 1-Shot} & \textbf{5-Way 5-Shot} \\
\hline
\multirow{5}{*}{A} & ProtoNet & 59.4 & 78.6 \\
 & ProtoNet+A & 61.4 & 80.7 \\
 & ProtoNet+B & 63.5 & 81.2 \\
 & ProtoNet+B+ & 64.9 & 81.9 \\
 & ProtoNet+A+B+ & \textbf{66.5} & \textbf{84.1} \\
\hline
\multirow{5}{*}{B} & ProtoNet & 50.3 & 66.9 \\
 & ProtoNet+A & 53.0 & 70.1 \\
 & ProtoNet+B & 53.8 & 70.6 \\
 & ProtoNet+B+ & 55.2 & 71.3 \\
 & ProtoNet+A+B+ & \textbf{59.3} & \textbf{72.5} \\
\hline
\multirow{5}{*}{C} & ProtoNet & 61.2 & 79.0 \\
 & ProtoNet+A & 63.5 & 81.5 \\
 & ProtoNet+B & 64.1 & 81.9 \\
 & ProtoNet+B+ & 66.0 & 83.2 \\
 & ProtoNet+A+B+ & \textbf{68.1} & \textbf{85.7} \\
\hline
\end{tabular}
\end{table}


As shown in Table~\ref{tab:ablation}, the proposed model achieves a significant improvement in classification accuracy over ProtoNet across all three environments. Notably, in Environment C, the model attains a classification accuracy of 68.1\% in the 5-way 1-shot setting and 85.7\% in the 5-way 5-shot setting, representing an increase of 6.9\% and 6.7\%, respectively, compared to the original ProtoNet.

Both the attention mechanism and data augmentation strategies contribute effectively to improving classification accuracy. For instance, in Environment A under the 5-way 1-shot setting, the baseline ProtoNet achieves an accuracy of only 59.4\%. The feature-level attention module enhances the representation of crucial feature dimensions by adjusting the Euclidean distance with attention scores, thereby mitigating feature sparsity issues. Consequently, incorporating this attention mechanism raises accuracy to 61.4\%, reflecting a 2.0\% improvement. 
Furthermore, applying a two-stage data augmentation approach increases accuracy to 63.5\%, yielding a 4.1\% improvement.
The progressive data augmentation strategy achieves even better results, enhancing accuracy by 5.4\% relative to ProtoNet. This improvement arises because progressive data augmentation involves additional training phases, albeit with a slightly longer training time. Ultimately, integrating both the attention mechanism and curriculum-inspired data augmentation yields an accuracy of 66.5\%, marking a substantial 7.1\% increase over ProtoNet.

Similarly, under the 5-way 5-shot setting, incorporating only the attention module results in an accuracy of 80.7\%, representing a 2.1\% improvement over ProtoNet. Applying the two-stage data augmentation strategy further increases accuracy by 2.6\%, while the progressive data augmentation approach yields a 3.3\% improvement over ProtoNet. Finally, integrating the attention-based prototype network with curriculum-inspired data augmentation leads to a final accuracy gain of 5.5\%.

These enhancements demonstrate a significant improvement in model performance, with the highest observed accuracy gains reaching 7.1\% (1-shot) and 5.5\% (5-shot). Furthermore, the experimental results indicate that, although the classification accuracy varies across different environments, the proposed model exhibits strong robustness and adaptability to diverse data collection scenarios.

\section{Conclusion}
\label{Conclusion}

In this paper, we proposed ProFi-Net, a novel few-shot gesture recognition model based on wireless signals, which combines a prototype-based feature attention mechanism with curriculum-inspired data augmentation. By leveraging a feature-level attention module, our approach refines the distance metric within the embedding space, effectively mitigating the issue of feature sparsity inherent in few-shot learning. Furthermore, the progressive data augmentation strategy applied on the query set gradually increases the difficulty of training samples, thereby enhancing model generalization and robustness. Extensive experimental results across multiple environments and various few-shot settings confirm that ProFi-Net significantly outperforms conventional prototype networks and other state-of-the-art models. Future work will focus on further optimizing the curriculum schedule and exploring additional temporal dynamics to further improve recognition performance in even more challenging real-world scenarios.

\bibliographystyle{IEEEtran}
\bibliography{IEEEabrv.bib,ref.bib}

\begin{thebibliography}{10}
\providecommand{\url}[1]{#1}
\csname url@samestyle\endcsname
\providecommand{\newblock}{\relax}
\providecommand{\bibinfo}[2]{#2}
\providecommand{\BIBentrySTDinterwordspacing}{\spaceskip=0pt\relax}
\providecommand{\BIBentryALTinterwordstretchfactor}{4}
\providecommand{\BIBentryALTinterwordspacing}{\spaceskip=\fontdimen2\font plus
\BIBentryALTinterwordstretchfactor\fontdimen3\font minus \fontdimen4\font\relax}
\providecommand{\BIBforeignlanguage}[2]{{%
\expandafter\ifx\csname l@#1\endcsname\relax
\typeout{** WARNING: IEEEtran.bst: No hyphenation pattern has been}%
\typeout{** loaded for the language `#1'. Using the pattern for}%
\typeout{** the default language instead.}%
\else
\language=\csname l@#1\endcsname
\fi
#2}}
\providecommand{\BIBdecl}{\relax}
\BIBdecl

\bibitem{hao2020wi}
Z.~Hao, Y.~Duan, X.~Dang, Y.~Liu, and D.~Zhang, ``Wi-sl: Contactless fine-grained gesture recognition uses channel state information,'' \emph{Sensors}, vol.~20, no.~14, p. 4025, 2020.

\bibitem{liu2020human}
X.~Liu, H.~Chen, A.~Montieri, and A.~Pescap{\`e}, ``Human behavior sensing: challenges and approaches,'' \emph{Journal of Ambient Intelligence and Humanized Computing}, vol.~11, no.~12, pp. 6043--6058, 2020.

\bibitem{botros2022day}
F.~S. Botros, A.~Phinyomark, and E.~J. Scheme, ``Day-to-day stability of wrist emg for wearable-based hand gesture recognition,'' \emph{IEEE Access}, vol.~10, pp. 125\,942--125\,954, 2022.

\bibitem{qi2020surface}
J.~Qi, G.~Jiang, G.~Li, Y.~Sun, and B.~Tao, ``Surface emg hand gesture recognition system based on pca and grnn,'' \emph{Neural Computing and Applications}, vol.~32, pp. 6343--6351, 2020.

\bibitem{shen2020gesture}
S.~Shen, K.~Gu, X.-R. Chen, C.-X. Lv, and R.-C. Wang, ``Gesture recognition through semg with wearable device based on deep learning,'' \emph{Mobile Networks and Applications}, vol.~25, pp. 2447--2458, 2020.

\bibitem{hussain2021comprehensive}
T.~Hussain, K.~Muhammad, W.~Ding, J.~Lloret, S.~W. Baik, and V.~H.~C. De~Albuquerque, ``A comprehensive survey of multi-view video summarization,'' \emph{Pattern Recognition}, vol. 109, p. 107567, 2021.

\bibitem{gao2020vision}
Y.~Gao, X.~Lu, J.~Sun, X.~Tao, X.~Huang, Y.~Yan, and J.~Liu, ``Vision-based hand gesture recognition for human-computer interaction-a survey,'' \emph{Wuhan University Journal of Natural Sciences}, vol.~25, no.~2, pp. 169--184, 2020.

\bibitem{peng2025stability}
X.~Peng, Y.~Hu, T.~Liu, Y.~Wu, T.~Saito, and T.~Toda, ``Stability-enhanced human activity recognition with a compact few-channel mm-wave fmcw radar,'' \emph{IEEE Transactions on Radar Systems}, 2025.

\bibitem{anguita2025multi}
M.~{\'A}. Anguita-Molina, P.~J. Cardoso, J.~M. Rodrigues, J.~Medina-Quero, and A.~Polo-Rodr{\'\i}guez, ``Multi-occupancy activity recognition based on deep learning models fusing uwb localisation heatmaps and nearby-sensor interaction,'' \emph{IEEE Internet of Things Journal}, 2025.

\bibitem{chen2019dynamic}
J.~Chen, F.~Li, H.~Chen, S.~Yang, and Y.~Wang, ``Dynamic gesture recognition using wireless signals with less disturbance,'' \emph{Personal and Ubiquitous Computing}, vol.~23, pp. 17--27, 2019.

\bibitem{gu2023wireless}
W.~Gu, S.~Yan, J.~Xiong, Y.~Li, Q.~Zhang, K.~Li, C.~Hou, and H.~Wang, ``Wireless smart gloves with ultra-stable and all-recyclable liquid metal-based sensing fibers for hand gesture recognition,'' \emph{Chemical Engineering Journal}, vol. 460, p. 141777, 2023.

\bibitem{sruthi2024handfi}
P.~Sruthi, S.~Satapathy, and S.~K. Udgata, ``Handfi: Wifi sensing based hand gesture recognition using channel state information,'' \emph{Procedia Computer Science}, vol. 235, pp. 426--435, 2024.

\bibitem{tian2018wicatch}
Z.~Tian, J.~Wang, X.~Yang, and M.~Zhou, ``Wicatch: A wi-fi based hand gesture recognition system,'' \emph{Ieee Access}, vol.~6, pp. 16\,911--16\,923, 2018.

\bibitem{li2020wihf}
C.~Li, M.~Liu, and Z.~Cao, ``Wihf: Gesture and user recognition with wifi,'' \emph{IEEE Transactions on Mobile Computing}, vol.~21, no.~2, pp. 757--768, 2020.

\bibitem{wang2018channel}
F.~Wang, W.~Gong, J.~Liu, and K.~Wu, ``Channel selective activity recognition with wifi: A deep learning approach exploring wideband information,'' \emph{IEEE Transactions on Network Science and Engineering}, vol.~7, no.~1, pp. 181--192, 2018.

\bibitem{meng2021wihgr}
W.~Meng, X.~Chen, W.~Cui, and J.~Guo, ``Wihgr: A robust wifi-based human gesture recognition system via sparse recovery and modified attention-based bgru,'' \emph{IEEE Internet of Things Journal}, vol.~9, no.~12, pp. 10\,272--10\,282, 2021.

\bibitem{bu2022deep}
Q.~Bu, G.~Yang, X.~Ming, T.~Zhang, J.~Feng, and J.~Zhang, ``Deep transfer learning for gesture recognition with wifi signals,'' \emph{Personal and Ubiquitous Computing}, pp. 1--12, 2022.

\bibitem{tang2021wifi}
Z.~Tang, Q.~Liu, M.~Wu, W.~Chen, and J.~Huang, ``Wifi csi gesture recognition based on parallel lstm-fcn deep space-time neural network,'' \emph{China Communications}, vol.~18, no.~3, pp. 205--215, 2021.

\bibitem{kabir2022csi}
M.~H. Kabir, M.~A. Hasan, and W.~Shin, ``Csi-deepnet: A lightweight deep convolutional neural network based hand gesture recognition system using wi-fi csi signal,'' \emph{IEEE Access}, vol.~10, pp. 114\,787--114\,801, 2022.

\bibitem{vinyals2016matching}
O.~Vinyals, C.~Blundell, T.~Lillicrap, D.~Wierstra \emph{et~al.}, ``Matching networks for one shot learning,'' \emph{Advances in neural information processing systems}, vol.~29, 2016.

\bibitem{sung2018learning}
F.~Sung, Y.~Yang, L.~Zhang, T.~Xiang, P.~H. Torr, and T.~M. Hospedales, ``Learning to compare: Relation network for few-shot learning,'' in \emph{Proceedings of the IEEE conference on computer vision and pattern recognition}, 2018, pp. 1199--1208.

\bibitem{zhang2020deepemd}
C.~Zhang, Y.~Cai, G.~Lin, and C.~Shen, ``Deepemd: Few-shot image classification with differentiable earth mover's distance and structured classifiers,'' in \emph{Proceedings of the IEEE/CVF conference on computer vision and pattern recognition}, 2020, pp. 12\,203--12\,213.

\bibitem{wertheimer2021few}
D.~Wertheimer, L.~Tang, and B.~Hariharan, ``Few-shot classification with feature map reconstruction networks,'' in \emph{Proceedings of the IEEE/CVF conference on computer vision and pattern recognition}, 2021, pp. 8012--8021.

\bibitem{he2022revisiting}
J.~He, R.~Hong, X.~Liu, M.~Xu, and Q.~Sun, ``Revisiting local descriptor for improved few-shot classification,'' \emph{ACM Transactions on Multimedia Computing, Communications, and Applications (TOMM)}, vol.~18, no.~2s, pp. 1--23, 2022.

\end{thebibliography}

\end{document}